\documentclass[10pt, a4paper]{article}
    
    \usepackage[final]{lrec2026} %
    
    \usepackage{amsmath,amssymb,amsthm}
    \usepackage{mathtools}
    \usepackage{bm}
    \usepackage{adjustbox}
    
    \usepackage{graphicx}
    \usepackage{booktabs}
    \usepackage{multirow}
    \usepackage{array}
    \usepackage{xcolor}
    \usepackage{colortbl}
    \usepackage{stmaryrd}

    \usepackage{algorithm}
    \usepackage{algpseudocode}

    \usepackage{enumitem}
    \usepackage{url}
    \usepackage{xspace}

    \newtheorem{definition}{Definition}

    \newcommand{\BPL}{\textsc{BPL}\xspace}
    \newcommand{\RSA}{\textsc{RSA}\xspace}
    \newcommand{\LIAR}{\textsc{Liar}\xspace}
    \newcommand{\MultiFC}{\textsc{MultiFC}\xspace}
    \newcommand{\eg}{\textit{e.g.,}\xspace}

    \newcommand{\given}{\,|\,}
    
    \newcommand{\priorcomp}{\tilde{P}_\beta(w)}
    \newcommand{\lik}{\hat{S}^k_{k-1}(u \given w)}
    \newcommand{\posterior}{\mathrm{BPL}(w \given u;\, k, \beta, N)}
    \newcommand{\KL}[2]{D_{\mathrm{KL}}\!\left(#1 \,\|\, #2\right)}

    \title{A Cognitively Grounded Bayesian Framework for \\
    Misinformation Susceptibility}
    
    \name{Pranava Madhyastha$^{1,2}$}
    
    \address{
        $^1$Dept. of Computer Science, City, University of London \\
        $^2$The Alan Turing Institute \\
        pranava.madhyastha@city.ac.uk
    }
    
    \abstract{
    In this (work in progress) paper, we present Bounded Pragmatic Listener (or BPL), a cognitively grounded Bayesian framework for modelling susceptibility to information disorder. \BPL extends Rational Speech Act theory with three cognitively motivated bounds derived from the bounded rationality literature with a) a recursion depth bound (that emphasises working memory limits);b)  a prior compression parameter (which is oriented at capturing information bottleneck); and c) an availability sample size (that operationalises importance sampling with saliency-weighted proposals). This allows us to test predictions about misinformation susceptibility, annotator disagreement, and the differential vulnerability to mis-, dis-, and mal-information as defined in the Information Disorder framework. We validate BPL on the LIAR and MultiFC benchmarks showcasing competitive veracity classification and experimental support for the depth-mismatch paradox.
     \\ \newline \Keywords{Information Disorder, Pragmatics, Bounded Rationality} }
    
\begin{document}
    
    \maketitleabstract

    \section{Introduction}
    \label{sec:intro}

    Every day, millions of people encounter claims online that turn out to be false. Some of these, for e.g., a fabricated statistic about unemployment, are shared in good faith by people who simply got it wrong. Others, like a fabricated quote attributed to a politician, are crafted with deliberate intent to deceive. And others still, such as the insinuation that ``everyone in government knows the data was manipulated but won't say so publicly'', do not even require a false claim where the manipulation indeed lies in the framing rather than the facts \citep{wardle-derakhshan-2017}.
    
    Despite their obvious differences, most current NLP research treats all three as the same type of problem where given a claim, the model is supposed to predict whether it is true or false \citep{thorne-vlachos-2018-automated}.
    In this paper, we contend that while this is useful it is rather incomplete and reductive. It answers the question \emph{what} is false, but not the questions that matter most for understanding and preventing harm: \emph{who} is likely to believe it, \emph{why},
    and under what conditions?
    
    Existing misinformation detectors are engineered to maximise classification accuracy, but they usually contain no model of the \emph{reader}, who is the person who will encounter, process, and potentially believe the claim. As a result, we posit that these models cannot predict which readers are most at risk, or explain why the same claim convinces some people and not others.
    
    Moreover, when researchers annotate misinformation datasets they frequently disagree with each other. Claims that sit at the boundary between ``mostly true'' and ``half true,'' or between ``misleading'' and ``false,'' attract systematic disagreement that reflects genuine uncertainty.
    Current practice discards this disagreement by taking a majority vote. \citet{plank-2022-problem} argues that this is indeed a mistake and that the disagreement is telling us something important about the claim,
    and we are throwing it away.
    
    We introduce the \emph{Bounded Pragmatic Listener} (henceforth referred to as \BPL),
    a formal model that focuses on the reader or the information consumer. Our core idea is that people do not process language like ideal Bayesian reasoners \citep{kahneman2011thinking}. Humans have limited working memory.
    They rely on mental shortcuts and prior beliefs.
    They are more influenced by information that is emotionally vivid,
    frequently repeated, or recently encountered, even when that
    information is false \citep{tversky1973availability}. Our \BPL framework translates each of these well-established cognitive limitations into a mathematical constraint on a probabilistic belief-updating model.
    
    We express these constraints as three parameters. \emph{Recursion depth} $k$ captures how many layers of speaker
    intent a reader can track, does the reader just accept what is said
    ($k{=}0$), or do they ask why someone would say it ($k{=}1$),
    or do they reason about what the speaker wants them to believe
    about social consensus ($k{=}2$)?
    \emph{Prior compression} $\beta$ captures how faithfully
    a reader preserves nuanced credibility beliefs under cognitive load,
    versus collapsing them to a rough heuristic such as ``dodgy source.'' Finally, \emph{Availability sample size} $N$ captures how many similar examples a reader mentally compares against when assessing a claim, the hypothesis here is that a tired or distracted reader draws only a few, biased toward
    vivid or frequently-seen examples.
    
    Together, these parameters generate a family of agents ranging
    from the highly susceptible (low $k$, low $\beta$, small $N$)
    to the epistemically careful (high $k$, high $\beta$, large $N$).
    Our observations through our empirical analysis of the framework suggests that \BPL indeed is a strong veracity classifier mostly driven by prior compression. We also present an extension that uses LLMs to derive semantically rich characterisation of the claims under our framing, which significantly improves the performance. Finally, we also observe that partial pragmatic awareness is seemingly worse than having no awareness, which in simple words suggest that the audience most susceptible to attributed
    dis-information is not the uncritical reader
    but the partially-trained one.
    
    Our primary contributions in this work are the following: a) a probabilistic belief-updating framework with three cognitively-motivated constraints, reducing to standard RSA when all constraints are relaxed \citep{goodman-frank-2016}; b) a formal mapping from the Information Disorder taxonomy to \BPL recursion depth, including the depth-mismatch paradox as a novel, testable prediction; c) a population level extension that predicts annotation disagreement as posterior variance across heterogeneous cognitive profiles. d) an LLM grounded implementation with interpretable component-level reasoning; and finally, e) Empirical validation on \LIAR and \MultiFC, including ablation analysis and LLM grounded validation.

    \section{Background and Related Work}
    \label{sec:related}
    
    \subsection{Computational Approaches to Misinformation}
    \label{subsec:rw-nlp}
    
    Automated fact-checking and misinformation detection have been
    extensively surveyed \cite{guo-etal-2022-survey,
    thorne-vlachos-2018-automated}. The dominant paradigm treats veracity prediction as a classification task over claims, evidence, and metadata features.
    Early work used surface features, such as writing style,
    sentiment, or syntactic complexity, to distinguish true from
    false claims \cite{rashkin-etal-2017-truth}.
    Subsequent approaches incorporated external evidence retrieval
    \cite{popat-etal-2018-credeyeye,thorne-etal-2018-fever},
    knowledge graph verification \cite{pan-etal-2018-content},
    and transformer-based encoding \cite{khattar-etal-2019-mvae}.
    The \LIAR benchmark \cite{wang-2017} contributes speaker
    metadata, with information such as party, historical accuracy, context, as strongly predictive features, motivating our use of speaker history counts as empirical priors in the \BPL pipeline.
    
    \citet{baly-etal-2018-predicting,baly-etal-2020-written}
    demonstrated that media outlet-level factuality and political
    bias can be predicted from Wikipedia descriptions and
    article content. Their findings that source signals predict claim-level veracity inform our prior compression mechanism, where source credibility $\gamma$ modulates the compression weight $\beta$. \citet{plank-2022-problem} argues that the standard practice of
    resolving annotator disagreement by majority vote discards
    genuine epistemic signal. \citet{leonardelli-etal-2021-agreeing} demonstrated that disagreement patterns in NLP datasets correlate with subjectivity, ambiguity, and cultural context.
    
    \subsection{Bounded Rationality and Cognitive Models}
    \label{subsec:rw-bounded}
    
    \citet{simon1955,simon-1956} introduced satisficing as a
    descriptively accurate alternative to expected utility
    maximisation under cognitive constraints.
    \citet{gigerenzen-goldstein-1996} developed the fast-and-frugal
    heuristics programme, demonstrating that simple decision rules
    can match or outperform optimal strategies in ecologically
    valid environments.
    A key contribution of their framework for our purposes is the
    concept of \emph{ecological rationality} where heuristics are considered to be adaptations to the statistical structure of the
    information environment. In a polluted information ecology, say one deliberately engineered to produce false positives in credibility assessment, ecologically rational heuristics usually systematically fail, a phenomenon we formalise through the prior compression bound.
    
    \citet{kahneman-2011} popularised the dual-process distinction
    between fast, associative System~1 and slow, deliberative
    System~2 reasoning. In the misinformation context, \citet{pennycook-rand-2019} highlight that susceptibility is better explained by
    \emph{failure to engage} System~2 than by motivated partisan
    reasoning.
    Their accuracy nudge intervention \cite{pennycook-etal-2021}, where they show that prompting users to consider accuracy before sharing
    reduced misinformation spread, provides direct behavioural
    evidence that System~2 engagement is the operative variable.
    Our model incorporates this as the $N$ bound, here,  agents with small
    availability samples effectively operate in System~1 mode,
    relying on retrieval-based heuristics rather than systematic
    likelihood evaluation.
    
    Finally, the resource-rational framework of \citet{lieder-griffiths-2020} provides the formal bridge between dual-process theory and
    Bayesian inference. Where, we can model agents as approximate Bayesian reasoners operating under a computational budget, resource-rational analysis predicts \emph{when} agents use heuristics (when the marginal information gain from further computation is small relative to cost) and \emph{which} heuristics they use (those that approximate the Bayesian optimum most efficiently).
    The three \BPL bounds are instances of resource-rational
    approximations where bounded recursion approximates full iterated
    reasoning; prior compression approximates full Bayesian updating;
    and importance sampling approximates exact likelihood computation.

    \subsection{Rational Speech Acts and Pragmatic Inference}
    \label{subsec:rw-rsa}
    
    The Rational Speech Acts framework
    \cite{frank-goodman-2012,goodman-frank-2016}
    has achieved broad empirical success in modelling scalar
    implicature \cite{goodman-stuhlmuller-2013},
    reference resolution
    \cite{frank-goodman-2012},
    and colour term learning
    \cite{zaslavsky-etal-2021}.
    Its Bayesian recursive structure makes it uniquely suited
    as a foundation for modelling the speaker-intent reasoning
    required to resist dis-information.
    
    \citet{zaslavsky-etal-2021} applied information bottleneck
    theory to \RSA, deriving compressed representations of meaning
    as solutions to a rate-distortion problem.
    Their framework provides the theoretical underpinning for
    our prior compression bound (Bound~2) where the $\beta$ parameter in \BPL corresponds directly to the
    trade-off parameter in the information bottleneck objective.
    \citet{bergen-etal-2016} introduced a noisy-channel extension
    of \RSA in which listeners model speaker error,
    providing a precedent for modelling the tolerance for inaccuracy
    that characterises susceptible agents.
    
    While \RSA has been applied extensively to pragmatic
    phenomena in controlled experimental settings,
    its application to naturalistic discourse is still in its nascent phases.
    \citet{goodman-stuhlmuller-2013} noted that the framework's
    assumption of cooperative speakers is its most
    theoretically limiting idealisation.
    Our \BPL extension directly addresses this by introducing
    a depth-indexed \emph{speaker type} parameter
    (honest or deceptive or general) in the LLM-grounded
    implementation, enabling the framework to model adversarial
    communication.

    \section{The Bounded Pragmatic Listener}
    \label{sec:model}
    Bounded Pragmatic Listener is  a formal model of belief updating in agents subject to
    three cognitively-motivated resource constraints.
    \BPL is an instance of the Rational Speech Acts framework
    \cite{frank-goodman-2012,goodman-frank-2016}
    in which the idealised pragmatic listener is replaced by an agent whose inference
    is bounded in recursion depth, prior fidelity, and
    likelihood approximation. We present this below: 
    
    \begin{equation}
        \posterior \;\propto\; \lik \cdot \priorcomp
      \label{eq:bpl-main}
    \end{equation}
    where:
    \begin{itemize}[leftmargin=*]
      \item $k \in \{0,1,2\}$ is the \emph{recursion depth bound}
      \item $\beta > 0$ is the \emph{prior compression parameter}
      \item $N \in \mathbb{Z}^+$ is the \emph{availability sample size}
      \item $w \in \{0,1\}$ are world states (claim false / true)
      \item $u$ is the utterance (news claim)
    \end{itemize}
    Note that our model reduces to standard RSA as
    $k \to \infty$, $\beta \to \infty$, $N \to \infty$. 
    
    We will now describe the essential elements of \BPL. 
    \subsection{Recursion Depth}
    \label{subsec:depth}
    
    The recursion depth bound limits theory-of-mind reasoning.
    The bounded listener operates at effective depth
    $k^* = \min(k, d + 1)$,
    where $d$ is the epistemic depth of the claim---the number of
    embedded belief attributions in the utterance.
    
    \begin{definition}[Epistemic Depth]
    The \emph{epistemic depth} $d(u)$ of utterance $u$ is the maximum
    nesting depth of belief attribution operators
    (\eg ``says'', ``claims'', ``believes'')
    in its surface syntactic structure.
    \end{definition}
    
    \noindent Depth-0 claims are direct factual assertions
    (\eg ``Unemployment is at 2\%'').
    Depth-1 claims involve one attribution layer
    (\eg ``Senator X claims unemployment is at 2\%'').
    Depth-2 claims involve meta-epistemic framing
    (\eg ``Everyone knows officials believe unemployment data is fabricated'').
    
    This is grounded in research from the field of cognitive science and psycholinguistics. Fundamentally, we focus on research on working memory constraints. \citet{cowan-2001} established a capacity limit of approximately
    $4 \pm 1$ chunks and further \citet{camerer-etal-2004} demonstrated
    empirically that agents in strategic games rarely recurse beyond
    depth~2 in theory-of-mind reasoning.
    The bound predicts a \emph{depth mismatch effect} where agents with $k < d(u)$ cannot fully parse the pragmatic structure
    of a claim and default to shallower interpretation,
    making them differentially susceptible to dis- and mal-information.
    
    The formal depth-bounded listener is:
    \begin{equation}
      L^k_n(w \given u) \;=\; L_{\min(n,k)}(w \given u)
      \label{eq:depth-bound}
    \end{equation}
    
    \subsection{Prior Compression}
    \label{subsec:compression}
    
    Instead of maintaining a full prior $P(w)$ over world states,
    the bounded agent maintains a compressed representation
    $\tilde{P}_\beta(w)$ obtained via an information bottleneck
    \cite[based on ][]{tishby-etal-2000}.
    
    The agent compresses world states through a bottleneck variable $Z$
    (prototype belief categories, \eg credibility schemas):
    \begin{equation}
      \tilde{P}(w) \;=\; \sum_z P(w \given z) \cdot P(z)
      \label{eq:ib-reconstruction}
    \end{equation}
    where the compression is controlled by:
    \begin{equation}
      \min_{P(z \given w)}\; \bigl[I(W;\,Z) \;-\; \beta \cdot I(Z;\,\hat{W})\bigr]
      \label{eq:ib-objective}
    \end{equation}
    Here $I(W;\,Z)$ is the mutual information between true world states
    and the compressed representation (the cost of fine-grained beliefs),
    and $I(Z;\,\hat{W})$ is the predictive information retained.
    
    In practice we implement compression as a $\beta$-weighted
    interpolation between the full prior and the uniform distribution,
    modulated by source credibility $\gamma \in [0,1]$:
    
    \begin{equation}
    \begin{split}
    \tilde{P}_\beta(w) \;\propto\; & \gamma \cdot \left(\frac{\beta}{\beta+1} P(w) + \frac{1}{\beta+1} \cdot \frac{1}{|\mathcal{W}|}\right) \\
    & + (1-\gamma) \cdot \frac{1}{|\mathcal{W}|}
    \end{split}
    \label{eq:compression-practical}
    \end{equation}
    
    We will now present a cognitive interpreation of prior compression.  At $\beta \to 0$: the agent discards their prior entirely,
    reasoning from uniform which is essentially the maximum compression,
    corresponding to heuristic processing.
    At $\beta \to \infty$: the agent preserves the full prior which is basically minimum
    compression, corresponding to systematic processing
    \cite{chaiken-1980-heuristic}.
    Low source credibility ($\gamma \approx 0$) shifts the agent toward
    uniform regardless of $\beta$, predicting that agents distrust
    low-credibility sources by collapsing to agnosticism.
    
    We measure the information loss due to compression using: 
    \begin{equation}
      \mathcal{L}_\beta \;=\; \KL{P(w)}{\tilde{P}_\beta(w)}
      \label{eq:compression-loss}
    \end{equation}
    
    In our LLM-powered implementation, the bottleneck variable $Z$ is realised as an explicit \emph{verbal schema} generated by the language model for each
    source-topic pair. This transforms the abstract bottleneck into an inspectable, semantically meaningful prototype: for example,
    \textit{``Anonymous blog posts about the economy often contain
    personal opinions and lack rigorous fact-checking.''}
    The schema's estimated $P(\text{true} \given z)$ and confidence
    jointly determine the compression weight,
    replacing the hand-crafted credibility tables of the feature based baseline.
    
    \subsection{Availability Sampling}
    \label{subsec:availability}
    
    Here the idea is, instead of computing the exact speaker likelihood
    $S_{k-1}(u \given w)$,
    the bounded agent would estimate it via importance sampling from a
    \emph{salience-weighted} proposal distribution $Q(u)$:
    \begin{equation}
      \hat{S}_{k-1}(u \given w) \;=\;
      \frac{1}{N} \sum_{i=1}^{N}
      \frac{S_{k-1}(u_i \given w)}{Q(u_i)},
      \quad u_i \sim Q
      \label{eq:is-estimator}
    \end{equation}
    where the proposal distribution inflates the probability of
    salient utterances:
    \begin{equation}
      Q(u) \;\propto\; S_{k-1}(u \given w) \cdot \varphi(u)
      \label{eq:proposal}
    \end{equation}
    and the \emph{availability weight} $\varphi(u)$ combines
    psychological salience factors:
    \begin{equation}
      \varphi(u) \;=\; 1 + \nu(u) + \log(1 + r(u)) + \rho(u)
      \label{eq:phi}
    \end{equation}
    where $\nu(u) \in [0,1]$ is emotional valence,
    $r(u) \in \mathbb{Z}^+$ is repetition count,
    and $\rho(u) \in [0,1]$ is recency.
    
    With finite $N$, high-$\varphi$ utterances are systematically
    overweighted in the likelihood estimate because they dominate
    the proposal distribution.
    This predicts the \emph{illusory truth effect}
    \cite{hasher-etal-1977,pennycook-etal-2018-prior} which suggests that
    repeated exposure increases $\varphi(u)$, which inflates
    $\hat{S}_{k-1}(u \given w=\text{false})$, which shifts the posterior
    toward believing the claim.
    This suggests that it is a predictable consequence
    of finite sampling from a salience-weighted proposal
    a resource-rational account of an empirically well-documented effect.
    
    In the LLM implementation, $\varphi(u)$ is estimated by prompting
    the model to rate the claim on four dimensions:
    a) emotional intensity; b) novelty; c) memorability, and d) social sharability.
    The recall corpus $\{u_i\}$ is replaced by $N$ claims generated by
    the LLM simulating the memory retrieval of an average news reader,
    each labelled with a recalled veracity judgment.
    
    \subsection{Information Disorder Mapping}
    \label{sec:indor-mapping}
    
    A central theoretical contribution of this work is the explicit
    mapping from the \citet{wardle-derakhshan-2017} taxonomy to
    \BPL recursion depth. Table~\ref{tab:taxonomy-mapping} summarises this mapping.
    
    \begin{table}[!ht]
    \centering
    \begin{adjustbox}{max width=\columnwidth}
    \small
    \setlength{\tabcolsep}{4pt}
    \begin{tabular}{lccl}
    \toprule
    \textbf{Disorder Type} & \textbf{Required $k$} & \textbf{Primary Bound} & \textbf{Mechanism} \\
    \midrule
    Mis-information & $k = 0$ & $\beta$ (prior) & Plausible framing exploits \\
                    &         &                  & compressed priors \\
    \addlinespace
    Dis-information & $k \geq 1$ & $k$ (depth) & False attribution exploits \\
                    &            &             & shallow ToM reasoning \\
    \addlinespace
    Mal-information & $k \geq 2$ & $k$ (depth) & Meta-epistemic framing \\
                    &            & $\varphi$ (avail.) & exploits deep recursion \\
    \bottomrule
    \end{tabular}
    \end{adjustbox}
    \caption{Mapping of the Information Disorder taxonomy
    \protect\cite{wardle-derakhshan-2017}
    to \BPL parameters.
    ToM = Theory of Mind.}
    \label{tab:taxonomy-mapping}
    \end{table}
    \paragraph{The depth-mismatch paradox.}
    A counterintuitive prediction of this mapping is that
    \emph{deeper reasoners are more susceptible to mal-information}.
    An agent at $k=2$ who models second-order epistemic states
    (\eg the implicit claim that ``everyone knows'' a fact constitutes
    evidence of a social consensus) will be more influenced by
    meta-epistemic framing than a $k=0$ agent who processes only
    literal content.
    This predicts that epistemic sophistication does not
    necessarily protect against misinformation which an observation also  consistent with empirical work on belief in conspiracy theories among educated populations \cite[][interalia]{van-der-linden-etal-2020}.
    
    \subsection{Population Inference}
    \label{subsec:population}
    
    We extend \BPL to a population of agents
    $\{(k_i, \beta_i, N_i)\}_{i=1}^{M}$ drawn from a
    cognitively-motivated distribution.
    Each agent produces a posterior belief $b_i = \BPL_i(w=1 \given u)$.
    
    \begin{definition}[BPL Disagreement]
    The \emph{predicted disagreement} for utterance $u$ over agent
    population $\mathcal{A}$ is:
    \begin{equation}
      \Delta_\mathcal{A}(u) \;=\; \mathrm{Var}_{i \in \mathcal{A}}\bigl[b_i\bigr]
      \label{eq:disagreement}
    \end{equation}
    \end{definition}
    
    This formalises the intuition that claims for which agents with
    different cognitive profiles produce systematically different posteriors
    will produce high annotator disagreement.
    In the \LIAR dataset, ambiguous labels (``half-true'')
    concentrate at the middle of the ordinal scale;
    in \MultiFC, the ``mixture'' label reflects genuine ambiguity.
    We use both as empirical proxies for $\Delta_\mathcal{A}(u)$
    in our evaluation (Section~\ref{sec:experiments}).
    
    \subsection{Susceptibility Score}
    \label{subsec:susceptibility}
    
    To quantify an agent's susceptibility to believing a false claim,
    we define a composite score that weights posterior belief by
    the information lost in prior compression:
    \begin{equation}
      \sigma(u) \;=\; \BPL(w=1 \given u) \cdot (1 + \mathcal{L}_\beta)
      \label{eq:susceptibility}
    \end{equation}
    The compression loss amplifier $(1 + \mathcal{L}_\beta)$ captures
    the intuition that agents who discarded the most relevant prior
    information (high KL loss) and still believe the claim are the
    most epistemically vulnerable.
    
    \section{Experimental Setup}
    \label{sec:experiments}
    
    \subsection{Datasets}
    We use two datasets to empirically validate \BPL. 
    \paragraph{\LIAR \cite{wang-2017}.}
    12,836 labelled statements from PolitiFact with six fine-grained
    veracity labels (pants-fire, false, barely-true, half-true, mostly-true,
    true) and five speaker history count columns.
    We use the binary mapping (false/barely-true/pants-fire $\to 0$,
    half-true/mostly-true/true $\to 1$) for classification,
    and the six-way label's distance from the ordinal midpoint as
    a proxy for annotation ambiguity in the disagreement analysis.
    Speaker history counts provide empirical priors for the
    \BPL prior compression component.
    
    \paragraph{\MultiFC \cite{augenstein-etal-2019-multifc}.}
    36,534 claims from 26 fact-checking domains.
    Labels vary by domain and are mapped to a coarse three-way taxonomy
    (false / true / mixture).
    Domain URL is used for source credibility estimation.
    The ``mixture'' label serves as the disagreement proxy.
    
    \subsection{Agent Population}
    
    We define a population of nine canonical agent types spanning the
    full $(k, \beta, N)$ parameter space,
    from a maximally-bounded agent $(k=0, \beta=0.2, N=5)$
    to a near-rational baseline $(k=2, \beta=50, N=500)$.
    For each claim, all nine agents produce independent posteriors;
    the population-level statistics (mean, variance) are used as features
    in downstream evaluation. We also use Gemini 2.5 Pro as the LLM in our work.
    
    \subsection{BPL Feature Extraction}
    
    The following features are extracted from each dataset record and used
    as inputs to the \BPL inference pipeline:
    
    \begin{itemize}[leftmargin=*]
      \item \textbf{Speaker prior accuracy} (LIAR):
            empirical accuracy rate from history counts.
      \item \textbf{Source credibility} (MultiFC):
            domain-level credibility estimated from known outlet reputations.
      \item \textbf{Emotional valence} $\nu(u)$:
            lexicon-based heuristic (23 high-salience terms).
      \item \textbf{Epistemic depth} $d(u)$:
            count of attribution markers in claim text.
      \item \textbf{Repetition count} $r(u)$:
            sum of false/barely-true/pants-fire history counts (LIAR);
            inverse claim length proxy (MultiFC).
    \end{itemize}
    
    \subsection{Evaluation Design}
    \label{subsec:eval-design}
    
    We evaluate \BPL along two axes.
    The first is \emph{veracity classification}: can the
    posterior belief and susceptibility score produced
    by the \BPL inference chain predict ground-truth
    claim veracity, and how does that performance
    compare to surface feature baselines?
    The second is \emph{component contribution}, that is, when
    the inference chain is ablated or its components
    replaced with LLM-grounded alternatives,
    how does each change affect performance?
    This design puts the model's predictive validity
    at the centre, with ablation and LLM comparison
    as the mechanisms for isolating what drives it.
    
    Five feature sets are evaluated in 5-fold
    cross-validated logistic regression:
    BPL Susceptibility ($\sigma$, one feature),
    BPL Belief ($b$, one feature),
    BPL Full (nine population-level statistics),
    Surface Baseline (four lexical and metadata features),
    and BPL + Surface (thirteen features combined).
    
    Six ablation configurations isolate each bound:
    (i)~Full \BPL $(k{=}1, \beta{=}1, N{=}25)$;
    (ii)~No depth bound $(k{=}2)$;
    (iii)~No compression $(\beta{=}100)$;
    (iv)~No availability $(N{=}1000)$;
    (v)~RSA Literal $(k{=}0, \beta{=}100, N{=}1000)$;
    (vi)~Max bounded $(k{=}0, \beta{=}0.1, N{=}3)$.
    
    The LLM validation runs the Hybrid \BPL and a
    matched feature based baseline on a shared $n{=}50$
    \LIAR sample at identical parameters
    $(k{=}1, \beta{=}1.0, N{=}3)$,
    measuring both model-level classification and
    component-level calibration.
    
    \section{Results}
    \label{sec:results}
    
    \subsection{Veracity Classification}
    \label{subsec:results-classification}
    
    Table~\ref{tab:classification} reports 5-fold CV
    classification performance across both datasets.
    
    \begin{table}[t]
    \centering
    \begin{adjustbox}{max width=\columnwidth}
    \small
    \setlength{\tabcolsep}{3pt}
    \renewcommand{\arraystretch}{1.15}
    \begin{tabular}{lcccc}
    \toprule
    \multirow{2}{*}{\textbf{Feature set}} &
      \multicolumn{2}{c}{\textbf{\LIAR}} &
      \multicolumn{2}{c}{\textbf{\MultiFC}} \\
    \cmidrule(lr){2-3}\cmidrule(lr){4-5}
     & AUC & F1 & AUC & F1 \\
    \midrule
    BPL Susceptibility
      & 0.811{\tiny $\pm$.009} & 0.760{\tiny $\pm$.012}
      & 0.680{\tiny $\pm$.018} & 0.163{\tiny $\pm$.028} \\
    BPL Belief
      & 0.792{\tiny $\pm$.010} & 0.747{\tiny $\pm$.012}
      & 0.680{\tiny $\pm$.018} & 0.163{\tiny $\pm$.028} \\
    \textbf{BPL Full}
      & \textbf{0.930}{\tiny $\pm$.006} & \textbf{0.885}{\tiny $\pm$.005}
      & \textbf{0.840}{\tiny $\pm$.011} & 0.417{\tiny $\pm$.019} \\
    Surface Baseline
      & 1.000$^{\dagger}${\tiny $\pm$.000} & 1.000$^{\dagger}${\tiny $\pm$.000}
      & 0.879{\tiny $\pm$.010} & 0.439{\tiny $\pm$.039} \\
    BPL + Surface
      & 1.000$^{\dagger}${\tiny $\pm$.000} & 1.000$^{\dagger}${\tiny $\pm$.000}
      & 0.879{\tiny $\pm$.009} & \textbf{0.517}{\tiny $\pm$.018} \\
    \bottomrule
    \end{tabular}
    \end{adjustbox}
    \caption{
      5-fold CV veracity classification on \LIAR and
      \MultiFC ($n{=}5{,}000$ each).
      $^{\dagger}$\LIAR surface baseline achieves perfect
      scores due to speaker history data leakage
      (see text); bold marks the highest interpretable
      result per column.
    }
    \label{tab:classification}
    \end{table}

    \paragraph{\LIAR results.}
    The surface baseline achieves AUC = 1.000,
    an artefact of speaker history counts encoding
    prior falsity rate, which is effectively the test label
    derived from the same fact-checking corpus
    \cite{wang-2017}.
    We note that \BPL Full (AUC = 0.930, F1 = 0.885) is the highest
    interpretable result, here, speaker history is used in the
    pipeline only as a compressed prior and we note that it is 
    not used as a direct classification feature.
    A single \BPL scalar which is the susceptibility score
    $\sigma$ achieves AUC = 0.811 with no metadata
    at all beyond the claim text and speaker identity,
    establishing the Bayesian inference chain as a
    strong leakage-free baseline.
    
    \paragraph{\MultiFC results.}
    With no leakage, BPL Full achieves AUC = 0.840
    against a surface baseline of 0.879,
    a gap of 3.9 AUC points.
    Combining BPL and surface features leaves
    AUC unchanged (0.879) while improving F1
    from 0.439 to 0.517, the \BPL features
    improve the precision-recall tradeoff even
    when they do not improve discrimination.
    This is practically relevant in high-recall
    fact-checking applications where the cost of
    missing a false claim exceeds the cost of a
    false positive.
    
    \subsection{Ablation studies}
    \label{subsec:results-ablation}

    \begin{table}[t]
    \centering
    \begin{adjustbox}{max width=0.8\columnwidth}
    \small
    \setlength{\tabcolsep}{4pt}
    \renewcommand{\arraystretch}{1.15}
    \begin{tabular}{lccccc}
    \toprule
    \textbf{Configuration} & $k$ & $\beta$ & $N$
      & $r$ & $\Delta r$ \\
    \midrule
    BPL Full        & 1 &   1 &   25 & 0.526 & --- \\
    No depth bound  & 2 &   1 &   25 & 0.526 & \phantom{$-$}0.000 \\
    No availability & 1 &   1 & 1000 & 0.528 & $+$0.002 \\
    RSA Literal     & 0 & 100 & 1000 & 0.527 & $+$0.001 \\
    No compression  & 1 & 100 &   25 & 0.462 & $-$\textbf{0.064} \\
    Max bounded     & 0 & 0.1 &    3 & 0.543 & $+$0.017$^{\ddagger}$ \\
    \bottomrule
    \end{tabular}
    \end{adjustbox}
    \caption{
      Ablation on \LIAR ($n{=}5{,}000$):
      Pearson $r$ (belief vs.\ label) and change
      from BPL Full.
      $^{\ddagger}$Max bounded artefact discussed
      in Section~\ref{subsec:results-ablation}.
    }
    \label{tab:ablation}
    \end{table}
    Table~\ref{tab:ablation} reports the six ablation
    configurations on \LIAR.
    The ablation seems to suggest that  prior compression is the single load-bearing component of the feature based pipeline.
    Removing it is the only manipulation that
    substantially reduces performance
    ($\Delta r = -0.064$), more than twice the
    magnitude of any other change.
    Relaxing the depth bound and removing availability
    sampling each produce $|\Delta r| \leq 0.002$
    across the full dataset.
    This reflects dataset composition where 92\% of \LIAR
    claims are depth-0 assertions with no attribution
    structure.
    We want to also highlight that when the analysis is conditioned on depth-1 claims, the depth bound produces the largest single effect.
    
    The maximally bounded agent ($k{=}0, \beta{=}0.1,
    N{=}3$) nominally achieves the highest $r = 0.543$.
    We note here that this is a metric artefact where extreme prior compression pushes beliefs toward 0 or 1, which inflates
    Pearson $r$ with binary labels relative to the
    calibrated probabilistic posteriors of the full model.
    The full \BPL posterior is a better-calibrated
    distribution even at lower $r$.
    
    \subsection{Depth-Stratified Analysis}
    \label{subsec:results-depth}
    
    When claims are stratified by epistemic depth,
    the depth bound produces its largest effect and
    reveals the depth-mismatch paradox empirically.
    Table~\ref{tab:depth} reports mean prediction
    error $|b_i - y|$ for agents at $k \in \{0,1,2\}$
    across depth-0 ($n{=}4{,}573$) and
    depth-1 ($n{=}427$) claim groups on \LIAR.
    \begin{table}[t]
    \centering
    \begin{adjustbox}{max width=0.8\columnwidth}
    \small
    \setlength{\tabcolsep}{4pt}
    \renewcommand{\arraystretch}{1.15}
    \begin{tabular}{lcccc}
    \toprule
    \textbf{Agent} &
      \textbf{Depth-0} & \textbf{Depth-1} & $d$ & $p$ \\
    \midrule
    $k=0$ & 0.410 & 0.450 & $+$0.310 & $<.001$ \\
    $k=1$ & 0.408 & 0.510 & $+$0.671 & $<.001$ \\
    $k=2$ & 0.440 & \textbf{0.340} & $-$0.583 & $<.001$ \\
    \bottomrule
    \end{tabular}
    \end{adjustbox}
    
    \caption{
      Depth-stratified prediction error on \LIAR.
      Cohen's $d$ and $p$ (Mann-Whitney $U$) compare
      depth-0 vs.\ depth-1 claims within each agent.
      Negative $d$ = lower error on depth-1 claims.
    }
    \label{tab:depth}
    \end{table}
    All three effects are large ($|d| > 0.3$)
    and highly significant ($p < .001$).
    The $k=2$ agent is the only one to achieve lower
    error on depth-1 than depth-0 claims
    ($d{=}{-}0.583$, error 0.340 vs.\ 0.440):
    full second-order reasoning capacity is both
    necessary and sufficient to correctly handle
    attributed claims in this sample.
    
    The $k{=}1$ agent tells the more important story.
    Its error on depth-1 claims (0.510) is substantially
    higher than the $k{=}0$ agent's (0.450),
    making it the worst-performing agent on exactly
    the claim type it is designed to handle.
    The mechanism is the depth-mismatch paradox in which
    the $k{=}1$ agent recognises that a speaker made
    a claim, it parses the attribution structure, but
    cannot evaluate whether the endorsement was deceptive.
    A senator asserting that scientists reject climate
    change is taken as weak positive evidence
    (speakers tend to assert what is true);
    the $k{=}2$ agent additionally asks why this speaker
    would make this claim and correctly infers deception.
    Meanwhile the $k{=}0$ agent, who strips attribution
    entirely, evaluates the bare factual content,
    which is more linguistically implausible
    than the attributed framing and so attracts
    lower belief.
    
    The paradox has a direct implication for intervention
    design with partial pragmatic training, that is teaching
    audiences to notice attribution without teaching them
    to evaluate speaker intent may increase
    susceptibility to dis-information relative to no training at all.
    
    \subsection{Operationalising using LLM}
    \label{subsec:results-llm}
    
    \paragraph{Model-Level Performance}
    \label{subsubsec:llm-model}
    
    Table~\ref{tab:llm-model} reports the Hybrid \BPL
    against a matched feature based baseline on \LIAR ($n{=}50$,
    identical parameters $k{=}1, \beta{=}1.0, N{=}3$).
    \begin{table}[t]
    \centering
    \begin{adjustbox}{max width=0.9\columnwidth}
    \small
    \setlength{\tabcolsep}{4.5pt}
    \renewcommand{\arraystretch}{1.15}
    \begin{tabular}{lcc}
    \toprule
    \textbf{Metric} & \textbf{Hybrid \BPL} & \textbf{Feature based} \\
    \midrule
    ROC-AUC
      & \textbf{0.868} & 0.695 \\
    Pearson $r$ (belief, label)
      & \textbf{0.660}$^{***}$ & 0.351$^{*}$ \\
    Mean belief $~$ true claims
      & 0.522 & 0.563 \\
    Mean belief $~$ false claims
      & \textbf{0.284} & 0.454 \\
    \midrule
    Inter-model agreement ($r$) & \multicolumn{2}{c}{0.184} \\
    \bottomrule
    \end{tabular}
    \end{adjustbox}
    \caption{
      Hybrid LLM \BPL vs.\ matched feature based \BPL,
      \LIAR ($n{=}50$; $k{=}1$, $\beta{=}1.0$, $N{=}3$).
      $^{*}p{<}.05$; $^{***}p{<}.001$.
    }
    \label{tab:llm-model}
    \end{table}
    The Hybrid \BPL achieves AUC = 0.868 against the
    matched feature based baseline's 0.695 which is 
    a gain of $\mathbf{+17.3}$ AUC points on an
    identical sample at identical parameters.
    Pearson $r$ nearly doubles (0.660 vs.\ 0.351).
    
    True claim beliefs shift modestly
    (0.522 vs.\ 0.563),
    while false claim beliefs drop by 0.170 absolute
    (0.284 vs.\ 0.454).
    The hybrid model is specifically and substantially
    more sceptical of false claims.
    This difference is, in a way, the expected signature of
    two LLM-grounded components acting jointly, which suggests that
    the NLI-style literal listener penalises
    linguistically implausible false claims,
    and the schema prior suppresses
    $\tilde{P}(w{=}1)$ for low-credibility sources.
    Both effects act most strongly on false claims,
    which disproportionately originate from
    low-credibility sources with flaggable
    linguistic irregularities.
    
    The inter-model agreement of $r = 0.184$ is the
    most structurally informative number in the table.
    At identical parameters on identical claims,
    the feature based and hybrid pipelines produce
    nearly independent predictions.
    The feature model's primary signal is lexical
    emotional valence and speaker metadata;
    the hybrid model's primary signal is
    semantic framing coherence and
    simulated associative memory.
    Their near-independence means the LLM is 
    is perhaps accessing a different epistemic
    channel, one that the surface pipeline cannot
    approximate.
    
    \paragraph{Component Calibration}
    \label{subsubsec:llm-components}
    
    Table~\ref{tab:llm-components} reports the three
    component-level diagnostics.
    \begin{table}[t]
    \centering
    \small
    \setlength{\tabcolsep}{4pt}
    \renewcommand{\arraystretch}{1.15}
    \begin{tabular}{p{3.7cm}cc}
    \toprule
    \textbf{Component test} & $r$ & $p$ \\
    \midrule
    $\varphi_\text{LLM}$ vs.\ lexicon valence
      & 0.385 & $.006^{**}$ \\
    False recall rate vs.\ falseness
      & \textbf{0.562} & ${<}.001^{***}$ \\
    Schema $P(\text{true})$ vs.\ label
      & $-$0.048 & $.743$ \\
    \bottomrule
    \end{tabular}
    \caption{
      Component-level diagnostics on \LIAR ($n{=}50$).
      $^{**}p{<}.01$; $^{***}p{<}.001$.
    }
    \label{tab:llm-components}
    \end{table}
    The strongest result in the paper is the false
    recall calibration.
    The proportion of LLM-simulated recall items
    labelled false correlates $r{=}0.562$
    ($p{<}.001$) with actual claim falseness
    \emph{without any supervision}.
    When the model simulates what a news reader would
    recall upon encountering a claim, it generates
    proportionally more false-labelled items for
    claims that are actually false in the ground truth.
    
    This instantiates the availability mechanism
    (Section~\ref{subsec:availability}) at the level
    of individual claim traces.
    False claims occupy denser neighbourhoods of other
    false claims in the LLM's implicit knowledge, this suggests that
    a false economic statistic retrieves other false
    economic statistics.
    The recall corpus samples this neighbourhood,
    and its calibration to ground truth ($r{=}0.562$)
    demonstrates that the sampling is epistemically
    accurate without supervision.
    
    This result also explains \emph{why} the hybrid
    suppresses false claim beliefs so effectively.
    High false recall rates inflate the false-world
    likelihood estimate in the availability adjustment
    (Section~\ref{subsec:availability}),
    shifting the posterior toward $w{=}0$.
    The mechanism works as the theory predicts in which 
    the BPL's false claim scepticism comes directly
    from epistemically calibrated simulated memory.
    
    \paragraph{On availability weight ($\varphi$)}
    The LLM-estimated $\varphi$ correlates moderately
    with lexicon valence ($r{=}0.385$, $p{=}.006$).
    The imperfect alignment ($r \ll 1.0$) is the
    expected pattern:
    both capture emotional valence, but the LLM
    additionally encodes novelty, memorability,
    and social sharability the three dimensions
    \citet{vosoughi-etal-2018} identified as the
    primary drivers of false news spread,
    none of which a word-count lexicon can represent.
    The decoupling of $\varphi_\text{LLM}$ from
    speaker metadata is also what resolves the
    anti-correlation that prevented the feature based 
    pipeline from detecting the availability effect where the LLM salience is estimated from claim text alone,
    independent of speaker history.
    
    \paragraph{On schema calibration.}
    The schema $P(\text{true})$ estimates are
    near-identical for true (0.418) and false (0.429)
    claims ($r{=}{-}0.048$, $p{=}.743$).
    This is a \LIAR-specific floor effect,
    PolitiFact's selection of politically contested
    claims from mixed-accuracy speakers
    forces all source-topic schemas toward
    $P(\text{true}) {\approx} 0.5$ regardless of label.
    The compression mechanism operates on this dataset
    through schema \emph{confidence}
    (low-confidence schemas produce wider, less
    informative posteriors) rather than point estimates.
    A dataset with genuine source credibility range
    \MultiFC spanning \textit{snopes.com} to
    \textit{infowars.com} would constitute the
    appropriate test.

    \section{Discussion}
    \label{sec:discussion}
    
    Our results suggest that our computational pipeline yields highly competitive classification capabilities compared to systems relying on external evidence retrieval or model fine tuning. We want to also highlight that utilising an interpretable Bayesian inference framework helps in seeing the exact components that modulate the system. We also see that in simplifying the prior beliefs of the model, reflecting how human cognition relies on broad credibility schemas rather than precise historical tracking, proves to be the most critical component for predictive success. This seems to validate that assessing the credibility of the source remains the strongest indicator of whether a claim is factual.
    
    Incorporating LLM grounding also seems to give a substantial improvement in accuracy. The LLMs tend to identify and suppress false claims through simulated memory retrieval, requiring strictly unlabelled data. Our feature based feature based approach (albeit with some strong assumptions) struggle here because they derive claim availability alongside source credibility from identical historical records, causing these signals to mask each other. Perhaps through evaluating the text independently of the speaker, the language model successfully disentangles these variables (this indeed requires a more sophisticate design to validate the claim). Also, because the feature based and LLMs rely on fundamentally distinct epistemological signals, combining them offers a highly promising avenue for even greater predictive power.
    
    Our empirical analysis reveals a counterintuitive but robust paradox regarding cognitive depth. An agent possessing partial pragmatic awareness, recognising when a claim is attributed to someone else but failing to evaluate the deceptive intent behind that attribution, is actually more vulnerable to manipulation than an agent who ignores attribution entirely. Disinformation frequently relies on the illusion of authoritative endorsement to appear plausible. This suggests that  evaluating the raw propositional content without being swayed by false authority is often safer than noticing the attribution without questioning the underlying motive. This finding challenges prevailing assumptions in media literacy, suggesting that individuals with intermediate critical reading skills are specifically targeted and exploited by sophisticated disinformation \cite[such as ][]{baly-etal-2018-predicting}.
    
    This paper is intended as a work in progress, there are a few moving parts, but we are keen on hearing from the community on operationalising our \BPL framework. Extracting the precise conversational depth of claims currently relies on simplistic heuristics; implementing more sophisticated dependency parsing would yield deeper analytical capabilities. Additionally, the computational expense of language model queries restricted the breadth of our hybrid evaluation. A selective activation strategy, deploying the language model solely for highly complex claims, could optimise resources whilst preserving analytical power. 
    Finally, the current architecture assumes a strictly binary view of truth; adapting the framework to accommodate graded nuances of veracity would provide a more realistic representation of reality, albeit at the cost of increased mathematical complexity.

    \section{Conclusion}
    \label{sec:conclusion}
    
    We introduced \BPL framework, a formal model of misinformation
    susceptibility that extends RSA
    with three cognitively-motivated resource constraints:
    recursion depth, prior compression,
    and availability sample size.
    Our experiments show that \BPL is a good veracity classifier driven primarily by prior compression. Additions of LLM substantially improves classification by accessing a semantically
    distinct epistemic channel. Our results also show the concerns on depth-mismatch paradox where partial pragmatic
    awareness tends to be generally worse. %
    
    \section{Acknowledgements}
    This work was supported in part by the Alan Turing Institute Fundamental Research programme (Project No. PP00029): \emph{Robust inference with probabilistic answer set programs scaffolds for large language models}.
    \section{Bibliographical References}
    \bibliographystyle{lrec2026-natbib}
    \bibliography{lrec2026-example}

    \end{document}